\documentclass{article}
\usepackage{spconf,amsmath,graphicx}
\usepackage{times}
\usepackage{epsfig}
\usepackage{subcaption}
\usepackage{amssymb}
\usepackage{color}
\usepackage{soul}
\usepackage[linesnumbered, ruled]{algorithm2e}
\usepackage{multirow}
\usepackage{booktabs}
\usepackage{subfloat}
\usepackage{caption}
\usepackage{float}
\usepackage{multirow}
\usepackage{tabularx}

\input{macros}
\title{Learning Visually Consistent Label Embeddings for Zero-Shot Learning}

\name{%
    Berkan Demirel$^{1,2}$, Ramazan Gokberk Cinbis$^2$, Nazli Ikizler-Cinbis$^3$\thanks{This  work  was  supported  in  part by TUBITAK Career Grant 116E445.}}

\address{%
    $^1$HAVELSAN Inc., $^2$Middle East Technical University, $^3$Hacettepe University}

\begin{document}
%
\maketitle
\begin{abstract}
In this work, we propose a zero-shot learning method to effectively model knowledge transfer between classes via jointly learning visually consistent word vectors and label embedding model in an end-to-end manner. The main idea is to project the vector space word vectors of attributes and classes into the visual space such that word representations of semantically related classes become more closer, and use the projected vectors in the proposed embedding model to identify unseen classes. We evaluate the proposed approach on two benchmark datasets and the experimental results show that our method yields significant improvements in recognition accuracy.
\end{abstract}

\section{Introduction}
\label{sec:intro}

With the surge of deep learning models, there is a high demand of large-scale datasets for training classification models 
over a large number of classes.
However, annotating such large-scale data is both highly costly and labor-intensive. Zero-shot learning
(ZSL)  emerges as a promising alternative in this regard. ZSL is a form of learning to handle classification when
the labelled training data is available for only some of the classes (called {\em seen classes}, \ie training classes), and,
not for the others (called {\em unseen classes}, \ie test classes). The basic philosophy of this technique is transferring knowledge from seen to
unseen classes by utilizing prior information from various sources such as textual descriptions of classes (\eg
\cite{zhang2016learning1, lei2015predicting, elhoseiny2013write}), embeddings of class names (\eg \cite{xian2016latent,
akata2015evaluation, al2016recovering}) or attribute-based class specifications (\eg \cite{zhang2016learning1, farhadi09cvpr,
luo2018zero, al2015transfer, xian2016latent, al2016recovering, Demirel_2017_ICCV}). Overall, the performance of a ZSL
method heavily depends on the prior information as it is the primary factor determining the limits of cross-class knowledge sharing and transfer.

In this study, we aim to increase the success of label-embedding based ZSL models by incorporating visually meaningful
word vectors for class embeddings. More specifically, the word embeddings of class names used in label embedding
techniques are typically derived from textual information in previous work~\cite{xian2016latent, akata2015evaluation,
al2016recovering}. These word vectors tend to capture only semantic relations, ignoring the visual
resemblances between the corresponding visual concepts. We argue that this may cause a considerable loss of information
for ZSL for object recognition. Instead, we propose to ground label embeddings on visually meaningful
word vectors proposed by \cite{Demirel_2017_ICCV}, which aims to transform word embeddings such that each class name
and the corresponding combination of attribute names attain a high degree of similarity. 
Unlike \cite{Demirel_2017_ICCV}, however, instead of relying directly on the attribute-to-class associations in the transformed
word embedding space, we construct our final ZSL model using the image-to-class associations measured by a label-embedding classifier.

\begin{figure*}
  \includegraphics[width=\textwidth]{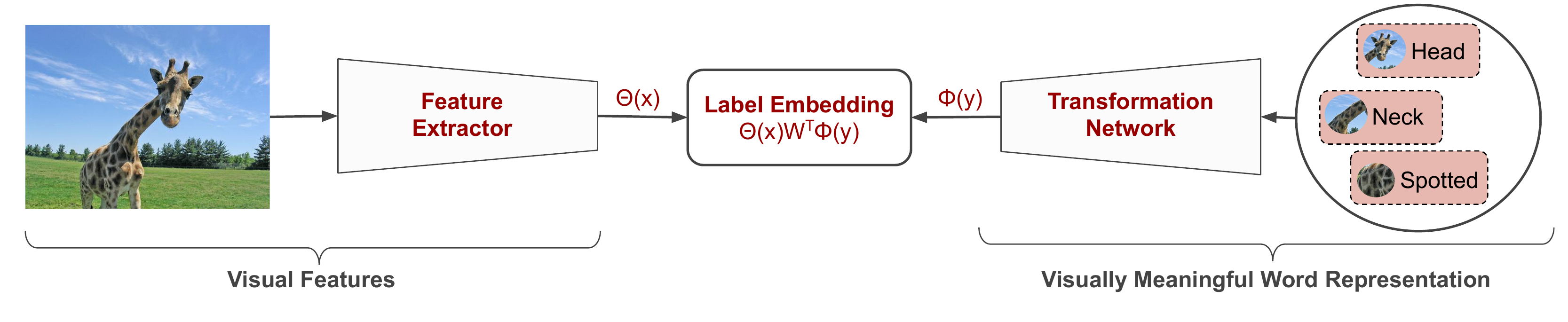}
  \caption{We propose a zero-shot learning approach based on visually meaningful word vectors and label embedding. In the transformation network,  visually more consistent word embeddings are learned. Then, the obtained visually consistent word embeddings are used in conjunction with the visual features in the label embedding model.}
  \label{main_figure}
\end{figure*}

In this work, we explore this idea and leverage visually meaningful word vectors as auxiliary data in label embedding to cover the bottlenecks of the both techniques. For this purpose, we learn visually more consistent word vectors and embedding space in an end-to-end manner by defining a joint loss function. This approach is illustrated in Figure~\ref{main_figure}. While using label embeddings, our approach utilizes the word representations transformed to a visually more consistent space. At test time, our zero-shot learning approach allows assigning novel images to unseen classes, purely based on class names.

To sum up, our main contribution in this work is utilizing the visually meaningful class name embeddings obtained by learning to associate 
corresponding attribute combinations and class names, and use them within a label embedding framework,
without requiring human-annotated attribute-class relations for the unseen classes.

In our experiments, we evaluate the proposed idea on two ZSL benchmark datasets, namely Animals with Attributes (AwA)~\cite{lampert13pami} and aPascal-aYahoo (aPaY)~\cite{farhadi09cvpr} datasets. We use word vectors which are obtained from GloVe method~\cite{pennington2014glove} to represent textual data. We also use CNN-M2K features~\cite{al2015transfer} to represent visual features and learn attribute based classifiers.
Our experimental results show that our method yields encouraging 
improvements in recognition accuracy on these benchmark datasets. 

\section{Related Work}

A large number of ZSL techniques have recently been proposed. The common goal of these techniques is information transfer between seen and unseen classes. Label embedding based approaches have been a particular focus of interest among the proposed techniques \cite{akata2013label,akata2015evaluation,norouzi2013zero,frome2013devise}. Akata \textit{et al.}~\cite{akata2013label} propose attribute label embedding (ALE) method that uses web scale annotation by image embedding technique~\cite{weston2010large} as an infrastructure. Different from WSABIE, ALE uses attributes as side information. Recently, Akata \textit{et al.}~\cite{akata2015evaluation} improve this ALE method by using embedding vectors that are obtained from huge text corpora instead of using human supervised class-attribute embedding. Xian~\textit{et al.} \cite{xian2016latent} use latent embedding spaces to encode comprehensive and discriminative visual characteristics of object classes. They learn a set of maps instead of learning a single bilinear map.

Attribute-based definitions of classes are one of the most important sources for transfering knowledge from seen to unseen classes. Lampert \textit{et al.}~\cite{lampert2009learning,lampert13pami} show that using attributes provide convenient and cost effective knowledge transfer between seen and unseen classes. Demirel \textit{et al.}~\cite{Demirel_2017_ICCV} use attribute information to learn visually more meaningful word representations. Unlike most other attribute based approaches, their method does not require the human supervised attribute-class relations at test time. Zhang \textit{et al.} \cite{zhang2016learning1} use multiple semantic modalities such as attributes and sentence descriptions. Moreover, they use visual space as the embedding space instead of using semantic space for embedding purpose. Ye and Guo~\cite{ye2017zero} propose an approach to find common high-level semantic components across the source and target domains where they have disjoint label spaces. For this purpose, they use discriminative sparse non-negative matrix factorization.

Al-Halah \textit{et al.}~\cite{al2015transfer} apply attribute label propagation on object classes, such that attributes are used at different abstraction levels, making it suitable for fine-grained zero-shot classification problem. In addition, 
several other zero-shot learning approaches have been developed based on attributes, classes, features and relations between them. Examples include those based on domain adaptation~\cite{deutsch2017zero}, semantic class prototype graph~\cite{fu2018zero}, unbiased embedding space~\cite{song2018transductive}, class prototype rectification~\cite{luo2018zero} or knowledge graphs~\cite{wang2018zero}. Recently, a few approaches have been proposed to use generative models for zero-shot learning~\cite{felix2018multi, verma2018generalized, long2018zero, zhang2019adversarial}.  


\section{Method}

In this paper, we build an end-to-end framework based on visually consistent word vectors and label embeddings. Basically, our method learns a transformation network that maps word vectors to an embedding space more suitable for zero-shot learning. For this purpose, we propose a  framework to jointly learn the word embedding transformation and the label embedding models in an end-to-end manner. 


In the rest of the section, we present the details of our approach. We first give a brief summary of the approach of \cite{Demirel_2017_ICCV}, which we utilize for learning visually meaningful word vectors, and then describe how we use these vectors as the side information in the label embedding model.

\subsection{Visually Meaningful Vector Space Word Vectors}

The introduction of distributed word vector representations, such as Word2Vec~\cite{mikolov2013distributed} or GloVe~\cite{pennington2014glove}, has been a step forward in semantic word representations, since these representations tend to capture the semantic nuances and relations between words more accurately. Based on their success, these vector space representations have witnessed a great attention in ZSL techniques and a large variety of other applications ranging from document retrieval to question answering. Nevertheless, in computer vision problems, semantic similarities at the word level may not be enough to model all the variances of the visual categories. For example, semantically similar words, such \textit{"wolf"} and \textit{"bear"} are not particularly close in visual domain, whereas visually consistent words such as \textit{"mole"} and \textit{"mouse"} can be far apart in semantic word domain. In order to account for such differences, \cite{Demirel_2017_ICCV} propose to learn a transformation on the word vectors that allows ZSL by comparing the pooled embeddings of attribute names and class names. Below we provide only a brief summary of the {\em image-based training} formulation of this approach, a more through explanation can be found in \cite{Demirel_2017_ICCV}.


In this formulation, the similarity between the class $y_i$ of an image $x_i$ and the set of associated attributes recognized in it should be higher than its similarity when another class embedding is used ($y_j$):
\begin{align} 
    && s(x_i,y_i) \geq s(x_i,y_j)+\Delta(y_i,y_j), && \forall{y_j \neq y_i}
\label{eq:ibt_constraint}
\end{align} 

\noindent where $\Delta$ is a margin function, indicates pairwise discrepancy value for each given training classes. In this inequality, $s(x,y)$ 
represents a compatibility function that measures the relevance between a pair of class and a set of posterior-probability weighted attributes,
formulated through a multilayer perceptron network. It also corresponds to a mapping that allows the transformation of word vectors from semantic to visually meaningful space. This approach is formalized as a constrained optimization problem:
%
\begin{align*}
    \min_{\Phi,\xi} \lambda ||\Phi||^2_2 + \sum_{i=1}^N \sum_{y_j  \neq  y_i} \xi_{ij} & \\
%
%
    s(x_i,y_i) \geq s(x_i,y_j) + \Delta(y_i,y_j) - \xi_{ij} & \hspace{5mm}  \forall y_j \neq y_i, \forall i
\end{align*}

\noindent where $\lambda$ is the regularization weight. Here, we learn a transformation matrix, which then we will call as $\Phi$. We refer to this transformation network as A2CN. 

\subsection{Label Embedding}

In order to use the visually consistent word vectors with visual data for ZSL, we prepare an embedding method:
\begin{equation}
f(x,y;W) = \Theta(x)^{T}W \Phi(y)
\label{eq:label_emb}
\end{equation}
\noindent where visual descriptors are denoted by $\Theta(x)$ and textual side information is denoted by $\Phi(y)$.
Moreover, $W$ matrix encodes textual and visual data to assign unseen test classes to correct class labels. This matrix
is designed as a dense layer in the multilayer perceptron network. In the Eq.~\ref{eq:label_emb}, $\Phi(y)$ side
information are feeded from A2CN model outputs. Cross-entropy loss is used to learn a proper embedding space. Softmax
classifier is also applied to normalise network predictions so that results can be interpreted as probabilities. 
Finally, we use the following joint loss function 
to learn transformation and embedding networks in an end-to-end manner. The final learning formulation, therefore, takes
the following form:
%
\begin{align*}
%
    \min_{\Phi,\xi,W} \sum_{i=1}^N \sum_{y_j  \neq  y_i} \xi_{ij} -\sum_{i=1}^N \log \frac{ \exp \Theta(x_i)^{T}W \Phi(y_i) }{ \sum_{y^\prime \neq y_i} \exp \Theta(x_i)^{T}W \Phi(y^\prime) } \hspace{2mm}  \\
    s(x_i,y_i) \geq s(x_i,y_j) + \Delta(y_i,y_j) - \xi_{ij} \hspace{5mm} \forall y_j \neq y_i, \forall i
%
\label{eq:loss}\end{align*}
where $\ell_2$ regularization is additionally applied to the parameters $W$ and $\Phi$, but omitted from the equation for brevity.

\section{Experiments}

In this section, we present the details of our experiments. First, we give a brief information about the datasets, and then explain initial word embeddings and visual features. Then, we give the details of our experiments, where we compare the proposed approach with its unsupervised and supervised counterparts.

\subsection{Datasets}
To evaluate the proposed approach, we use two benchmark ZSL datasets, namely Animals with Attributes (AwA)\cite{lampert13pami} and aPascal-aYahoo (aPaY)\cite{farhadi09cvpr}. The AwA dataset consists of images with 50 animal classes, 40 of which are training and 10 of which are test. 85 per-class attributes are defined on these classes. aPaY dataset consists of images from two different sources. Training part is obtained from Pascal VOC 2008~\cite{pascal-voc-2008} dataset, containing 20 classes. The test part is collected using Yahoo search engine and it contains 12 classes; totaling up to 32 completely different classes overall. Images in aPaY dataset are annotated with 64 binary per-image attributes. We follow the same experimental setup as in \cite{Demirel_2017_ICCV} for AwA and aPaY experiments.

\subsection{Implementation Details}
\label{imp_details}
Initially, we use 300-dimensional word embedding vectors which are obtained from GloVe method as described in A2CN method~\cite{Demirel_2017_ICCV} for fair comparison. Following the A2CN method, we obtain word vectors for each class and attribute names. If attribute or class names consist of multiple words, word vectors are obtained for each word, then the average of these vectors is used.

For AwA and aPaY datasets, we utilize the CNN-M2K features~\cite{al2015transfer}, where images are resized to 256x256 and mean image subtraction is applied. Outputs of the last hidden layer are extracted for image representation, as also described in \cite{Demirel_2017_ICCV}.

We define our method as a three layer feed-forward network. We use 2-fold cross-validation to determine optimal number of hidden units. Adam optimizer \cite{kingma2014adam} is used for stochastic optimization and the learning rate value is set to 1e-4. The proposed joint loss function is used to learn transformation and embedding networks in an end-to-end manner.

\begin{figure*}
 {persian+cat ~~~~~hippo ~~~~~~~~leopard ~~~~~~~~~h.whale ~~~~~~~~~~~seal ~~~~~~chimpanzee ~~~~~rat ~~~~~~~~~~~~~g.panda ~~~~~~~~~~~pig ~~~~~~~~raccoon\par }
    \includegraphics[width=\textwidth,height=\textheight,keepaspectratio]{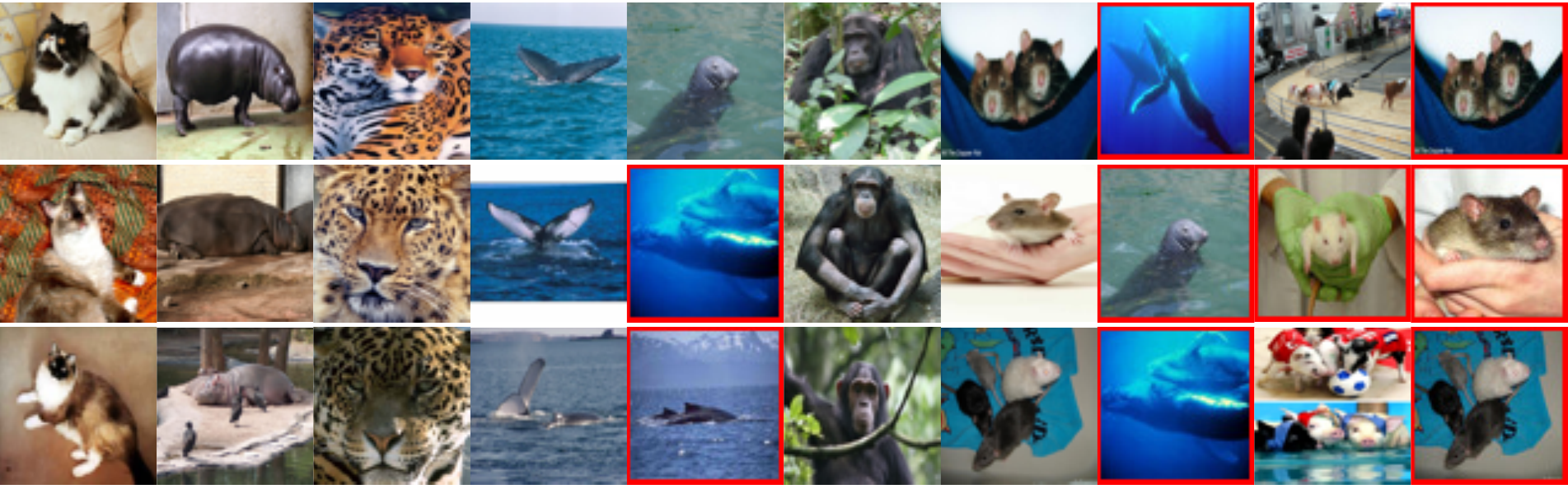}
    \caption{Top-3 highest scoring images using PBT method in the AwA dataset.}
    \label{result:awa_pbt}
\end{figure*}

\subsection{ Experimental Results}

In our experiments, we first evaluate our method using two different training methods, PBT and IBT, that are proposed by~\cite{Demirel_2017_ICCV} to handle ZSL problem. PBT stands for \textit{(Predicate-based Training)} and IBT stands for \textit{(Image-based Training)}. We measure the performance using the normalized per-class accuracy and the results are shown Table \ref{table:ourMethods}. According to the obtained results, it seems that our method provides a noticeable progress using PBT training method, where there is a $2.9\%$ accuracy increase in AwA and $4.9\%$ increase in aPaY datasets. For the IBT method, some improvements are also observed in the aPaY dataset, whereas the recognition performance slightly degrades on AwA dataset. This may be due to the fact that the parameters learned during the cross-validation may not produce the best results for the test classes. We also believe that PBT method is more important than IBT, because it only uses predicate matrix to learn meaningful word vectors, so it is a more generalizable method with less training data.

\begin{table}
\begin{center}
\caption{Zero-shot classification performance of proposed method on AwA and aPaY datasets.} 
\label{table:ourMethods} 
\begin{tabular}{ c c c c }
\hline
    Test&Method & AwA & aPaY \\
\hline
    \multirow{2}{6em}{PBT} 
    & A2CN\cite{Demirel_2017_ICCV}& 60.7 & 29.4\\
    & Our Method& \textbf{63.6} & \textbf{34.3}\\
    \midrule
    \multirow{2}{6em}{IBT}
    & A2CN\cite{Demirel_2017_ICCV} & \textbf{69.9} & 38.2 \\
    & Our Method& 68.6 & \textbf{40.8} \\
\hline
\end{tabular}
\vspace{-3mm}
\end{center}
\end{table}

We also compare our approach with various unsupervised and supervised counterparts presented in the literature. The results are shown on Table \ref{table:comparisonSOA}. Here, supervised methods require additional information about test classes such as class-attribute relations. Unsupervised ZSL methods do not require any human supervision about the unseen test classes. When we review the results on Table \ref{table:comparisonSOA}, we observe that our method obtains higher classification performance on aPaY dataset, compared to its unsupervised counterparts. On AwA dataset, it outperforms its unsupervised counterparts, except for A2CN\cite{Demirel_2017_ICCV} method. 

Our ZSL method also produces comparable results to some of the supervised counterparts. Another interesting direction to note is that, while high accuracies can potentially be obtained using the recently proposed data generation models~\cite{felix2018multi, verma2018generalized, long2018zero, zhang2019adversarial}, these works are orthogonal to proposed method, and, in principle, these techniques can be used in combination with the ZSL model proposed in this work. We plan to investigate this line of research in future work.

\begin{table}
\begin{center}
\caption{Comparison of the related ZSL literature. }
\label{table:comparisonSOA}
\begin{tabularx}{\linewidth}{ c l l l l l }
\hline
Test supervision & Method & AwA & aPaY \\
\hline
\multirow{8}{6em}{unsupervised} 
    &DeViSE\cite{frome2013devise}&44.5 & 25.5 \\
    &ConSE\cite{norouzi2013zero}&46.1 & 22.0 \\
    &Text2Visual\cite{elhoseiny2013write,bo2010twin}&55.3 & 30.2 \\
    &SynC\cite{changpinyo2016synthesized}&57.5&-\\
    &ALE\cite{akata2015evaluation} &58.8 & 33.3 \\
    &LatEm\cite{xian2016latent}&62.9 & - \\
    &CAAP\cite{al2016recovering}&67.5 & 37.0 \\
    &A2CN\cite{Demirel_2017_ICCV} & \textbf{69.9}& 38.2 \\
    &Our Method & 68.6& \textbf{40.8} \\
    \midrule
\multirow{6}{6em}{supervised} & 
    DAP\cite{lampert13pami} & 54.0& 28.5 \\
    & ENS\cite{rohrbach2011evaluating} & 57.4& 31.7 \\
    & HAT\cite{al2015transfer} & 63.1& 38.3 \\
    & ALE-attr\cite{akata2015evaluation} & 66.7& - \\
    & SSE-INT\cite{zhang2015zero} & 71.5& 44.2 \\
    & SynC-attr\cite{changpinyo2016synthesized}&76.3&-\\
\hline
\end{tabularx}
\end{center}
\vspace{-4mm}
\end{table}

Finally, Figure~\ref{result:awa_pbt} illustrates qualitative examples of the results of the our approach. In this figure, we show the top-3 scoring images produced by our proposed method on AwA dataset. The misclassified images are marked with red. According to this illustration, misclassifications tend to occur across visually similar classes, as expected.

\section{Conclusion}

The performance of the zero-shot learning approaches depend on the shared prior information between training and unseen test classes; therefore, it is very critical that the prior information is accurate, consistent and comprehensive. In this work, we have aimed to improve zero-shot recognition by using visually meaningful word vectors within the label embedding framework. The experimental results show the effectiveness of the proposed approach. 

\bibliographystyle{IEEEbib}
\small{\bibliography{paper}}

\end{document}